\documentclass[10pt,twocolumn,letterpaper]{article}

\usepackage[pagenumbers]{cvpr} 

\usepackage{multirow}
\usepackage{array}
\usepackage{xcolor} 

\usepackage{colortbl} 
\usepackage{pifont}   
\newcommand{\cmark}{\ding{51}}
\newcommand{\xmark}{\ding{55}}

\definecolor{cvprblue}{rgb}{0.21,0.49,0.74}
\usepackage[pagebackref,breaklinks,colorlinks,allcolors=cvprblue]{hyperref}

\def\confName{CVPR}
\def\confYear{2027}

\title{Lines Are Ideas: Visual Creativity Assessment in Hyperbolic Space}

\author{Baoliang Chen\\
City University of Hong Kong \\
Kowloon, Hong Kong SAR, China\\
{\tt\small}
\and
Zihao Lin\\
South China Normal University\\
Guangzhou, Guangdong, China\\
{\tt\small}
\and
Hao Liu\\
Mango TV (MGTV)\\
Changsha, Hunan, China\\
{\tt\small }
\and
Xiangjie Sui \thanks{Corresponding author.} \\
City University of Macau\\
Taipa, Macao SAR, China\\
{\tt\small }
}

\begin{document}
\maketitle
\begin{abstract}
Assessing human creativity from drawings provides a scalable way to evaluate human creative ability. However, existing approaches either rely on costly expert annotation or treat creativity as a single regression target, overlooking its underlying structure. Psychological theories have long emphasized that creativity emerges from balancing \textbf{novelty} and \textbf{appropriateness}. Motivated by this principle, we recast creativity as \emph{structured semantic deviation} between paired drawings, where novelty measures the extent of departure from the original base sketch, and appropriateness determines whether such departures remain meaningful. This formulation naturally leads to a hyperbolic representation, in which novelty is quantified by geodesic distance and appropriateness is evaluated conditionally via a learned decision boundary, further regulated by an abstraction-conditioned adaptive constraint. Experiments on AuDrA-Drawings demonstrate state-of-the-art alignment with expert judgments and strong robustness under rater and task shifts. Beyond performance gains, our framework reveals an interpretable radial organization of creative variation, offering a principled geometric foundation for automatic creativity assessment.
\end{abstract}    
\section{Introduction}

Creativity is central to human intelligence, yet remains difficult to assess objectively~\cite{Creativity1950,Long2022Dissecting}. Existing approaches often rely on subjective and time-consuming human evaluations~\cite{ttct,cat,tcia}, motivating the need for more intuitive and perception-grounded alternatives. Psychophysical evidence~\cite{finke1989reinterpreting,dynamic_creativity} suggests that creativity can manifest in remarkably simple settings: given the same sparse drawing composed of only a few lines, individuals may produce highly diverse reinterpretations, ranging from literal reproductions to abstract and imaginative transformations. Such variability enables creativity to be assessed by directly comparing an original sketch with its creative outcome. Following this vein, several psychophysics studies~\cite{bainbridge2022tutorial,patterson2024audra,cropley2025automated} collected original--creative image pairs with overall creativity scores, which subsequently inspired computational models for automatic creativity assessment~\cite{wang2023exploring,zhang_painting,panfilova_resnet,Pencils_to_Pixels}. 

\begin{figure}
    \centering
    \includegraphics[width=1\linewidth]{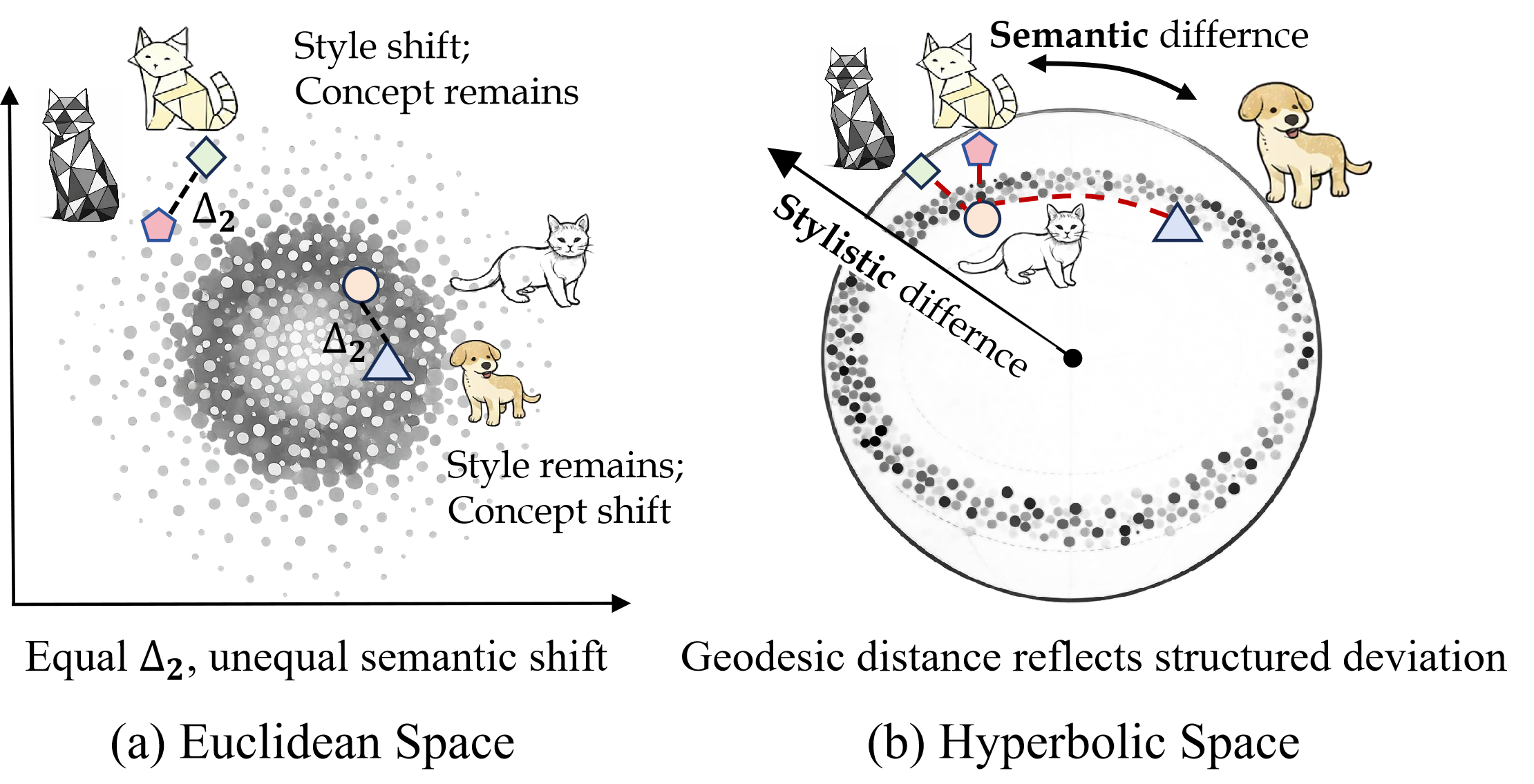}
    \caption{\textbf{Motivations}. (a) Euclidean geometry is struggles to model the non-uniform, hierarchical concept spaces; (b) hyperbolic geometry provides structure-aware representations via radius- and geodesic-based deviation.}
    \label{fig:intro}
\end{figure}
A critical advantage of the paired setting is that it turns creativity into a \textit{measurable transformation}: the creative outcome should deviate from the original one to exhibit {\textit{novelty}} (\ie, originality and unexpectedness), while preserving enough semantic interpretability to remain {\textit{appropriate}} within the context. This novelty--appropriateness view is consistent with classic theories of creativity~\cite{sternberg1999concept}. However, turning this principle into a robust computational metric is non-trivial. Most computational approaches implicitly assume that creative artifacts reside in a Euclidean latent space~\cite{creativity_survey,creativity_DL}, and measure deviation using Euclidean distances, \eg, $\Delta_2=\|\mathbf{x}-\mathbf{y\|}_2$. This assumption conflicts with the structure of conceptual knowledge: conceptual spaces are \textit{non-uniform}, with dense regions around prototypical concepts and sparse regions for rare or abstract realizations; moreover, conceptual organization is inherently \textit{hierarchical} (\eg, animal $\rightarrow$ mammal $\rightarrow$ cat), which Euclidean geometry with its polynomial volume growth struggles to represent. As a result, the same Euclidean displacement may correspond to qualitatively different conceptual shifts -- as shown in Fig.~\ref{fig:intro}, it may flip a dense prototype from \textit{cat} to \textit{dog}, but in a sparse abstract region it may only alter rendering style while preserving the concept.

In this work, we argue that \textit{hyperbolic geometry} provides a suitable inductive bias for creativity assessment: its negative curvature yields exponential volume growth, aligning naturally with hierarchical and non-uniform conceptual organization, and has been effective for modeling taxonomies and semantic hierarchies~\cite{poincare,ganea2018hyperbolic}. 
We argue that the creative visual concepts share this structure, with prototypical cores and increasingly diverse yet sparser realizations toward the periphery. 
Accordingly, we formulate creativity through novelty and appropriateness in hyperbolic space: novelty is measured as the hyperbolic geodesic distance between the original and creative embeddings, while appropriateness is determined by whether such deviation remains semantically coherent, conditioned on semantic uncertainty and implemented via a hyperbolic decision boundary. 
We then map this appropriateness signal to the final creativity score, since creativity should reward novelty only when it is semantically meaningful rather than arbitrary distortion.

Our contribution can be summarized as follows:
\begin{itemize}[leftmargin=*,nosep]
    \item \textbf{Hyperbolic-space creativity assessment.} We propose a hyperbolic creativity assessment framework that models creative transformation as \emph{structured semantic deviation} and operationalizes creativity through a novelty--appropriateness formulation.
    \item \textbf{Style-aware constrained hyperbolic representation.} We introduce an style-conditioned adaptive constraint in a Poincar\'e ball, enabling controlled deviation that accommodates stylistic variation while preventing semantically incoherent drift.
    \item \textbf{Hyperbolic novelty--appropriateness assessment.} We measure novelty via hyperbolic geodesic distance and estimate appropriateness by combining novelty with semantic uncertainty and a hyperbolic decision boundary, yielding a final creativity score that rewards novelty only when it is semantically coherent.
\end{itemize}


\section{Related Work}
\subsection{Psychological Foundations of Creativity Assessment}
Creativity is most commonly defined in psychology as the ability to produce outcomes that are both \emph{novel} and \emph{appropriate} to task constraints. This dual-criterion formulation has been consistently emphasized in influential theories, notably by Sternberg and Lubart~\cite{sternberg1999concept}, and has become the dominant conceptual basis for creativity assessment.
Visual creativity evaluations largely operationalize this definition through task-based drawing or imagery paradigms. The figural Torrance Tests of Creative Thinking (TTCT)~\cite{ttct} assess how individuals transform given shapes into meaningful drawings, scoring originality and task-relevant elaboration. Similarly, the Test of Creative Imagery Abilities (TCIA)~\cite{tcia} evaluates originality and the meaningful development of mental images. More recent tablet-based protocols such as Multi-Trial Creative Ideation (MTCI)~\cite{dynamic_creativity} further capture fine-grained drawing behaviors across multiple trials, but still evaluate creative products primarily along originality and appropriateness-related dimensions.
Beyond specific task designs, the Consensual Assessment Technique (CAT)~\cite{cat} formalizes creativity evaluation via expert agreement, where judgments implicitly reflect both originality and contextual suitability. Despite diverse experimental settings and scoring schemes, these paradigms consistently converge on novelty and appropriateness as the core evaluative dimensions, motivating our explicit two-factor formulation for image creativity assessment.

\subsection{Computational Modeling of Visual Creativity}

Recent studies explore automatic visual creativity assessment using deep networks trained to predict human ratings from drawings or paintings, motivated by divergent thinking theories. Early CNN-based models were developed for TTCT-style tasks~\cite{cropley_figural}, with later extensions using deeper backbones and explainability tools~\cite{panfilova_resnet, zhang_painting}. More scalable systems predict continuous creativity scores from large annotated datasets, such as AuDrA~\cite{patterson2024audra}, or combine stylistic descriptors with content embeddings to improve interpretability, as in Pencils-to-Pixels~\cite{Pencils_to_Pixels}. {This content--style perspective inspires our use of semantic category alignment and stylistic abstraction cues for creativity assessment.}These methods demonstrate that visual creativity can be learned from data, but mainly rely on global score regression in Euclidean latent spaces.


%
Despite progress, three limitations remain. First, Euclidean embeddings poorly capture hierarchical and unevenly distributed conceptual structures, where novel ideas lie in sparse peripheral regions. Second, generalization across different creative prompts and semantic contexts remains limited~\cite{patterson2024audra}. Third, visual deviation is often rewarded without explicit constraints on semantic or functional appropriateness, conflating novelty with arbitrary variation and neglecting a core criterion of creativity~\cite{sternberg1999concept}.

\begin{figure*}[htbp]
    \centering
    \includegraphics[width=0.86\linewidth]{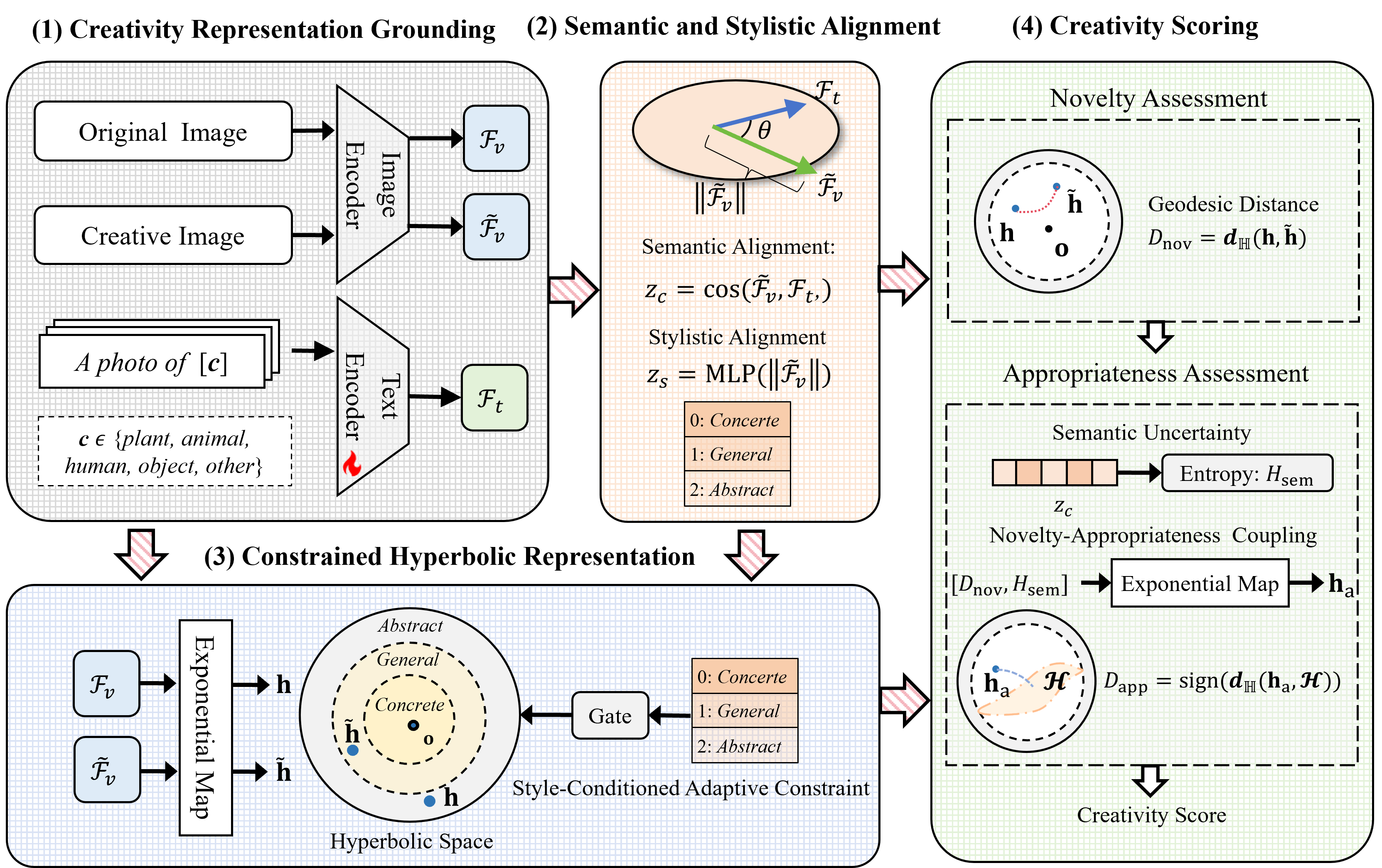}
    \caption{\textbf{Hyperbolic-space creativity assessment framework}.The pipeline consists of four stages: (1) multimodal feature extraction using CLIP encoders to obtain visual features $(\mathcal{F}_v,\widetilde{\mathcal{F}}_v)$ and semantic anchors $\mathcal{F}_t$; (2) semantic category alignment via cosine similarity to produce content logits $z_c$ and abstraction level alignment via an MLP head to obtain $z_s$; (3) constrained hyperbolic representation by exponential mapping into a Poincar\'e ball with an abstraction-conditioned adaptive constraint; and (4) creativity scoring via novelty $D_{\mathrm{nov}}=d_{\mathbb{H}}(\mathbf{h},\widetilde{\mathbf{h}})$ and appropriateness computed from $(D_{\mathrm{nov}}, H_{\mathrm{sem}})$, yielding the final creativity score.}
    \label{fig:framework}
    \vspace{-1em}
\end{figure*}

\section{Methodology}
Herein, we introduce the Hyperbolic-Space-based Creativity Assessment framework (denoted as HSCA).
Our core idea is to model creativity as a form of \emph{structured semantic deviation}: a creative image should exhibit sufficient novelty while remaining semantically interpretable. Accordingly, creativity is assessed as a conditional judgment in which appropriateness determines whether a novel deviation constitutes meaningful creativity. 
As shown in Fig.~\ref{fig:framework}, the framework proceeds in four stages: (1) multimodal feature extraction for obtaining content representations; (2) semantic and stylistic alignment to ground creative transformations; (3) constrained hyperbolic representation to regulate allowable deviation, and (4) novelty--appropriateness assessment to derive the final creativity score.

\subsection{Creativity Representation Grounding}
Creativity assessment over original-creative image pairs is inherently comparative, requiring a shared representational reference to quantify how creative image deviates from the original one. We therefore encode the original and creative images with a pretrained CLIP image encoder $\mathbf{E}_{\mathrm{v}}$, obtaining paired visual embeddings $\mathcal{F}_v, \widetilde{\mathcal{F}}_v \in \mathbb{R}^{D}$ in a shared visual-semantic space. To further ground the comparison in an interpretable semantic context, we introduce a small set of high-level content prompts: We construct prompts $\mathcal{T}(c)$ by filling the template ``\texttt{a photo of \{}$c$\texttt{\}}'', $c\in\mathcal{C}=$\{\textit{``plant'', ``animal'', ``human'', ``object'', ``other''}\}. These prompts serve as \emph{semantic anchors} that partition the embedding space into coarse conceptual regions. 
We then encode these prompts using the CLIP text encoder $\mathbf{E}_\mathrm{t}$:
\begin{equation}
\mathcal{F}_t = \mathbf{E}_\mathrm{t}\big( \mathcal{T}(c) \big
),
\quad \forall{c} \in \mathcal{C},
\end{equation}
where the $\mathcal{F}_t \in \mathbb{R}^{|\mathcal{C}| \times D}$ denotes the  textual embeddings.

\subsection{Semantic and Stylistic Alignment}
Following the content--style perspective ~\cite{Pencils_to_Pixels}, we extract two complementary alignment signals from each original and creative image pair to ground subsequent deviation modeling: a coarse \emph{semantic} alignment to high-level content anchors; and a \emph{stylistic} cue in terms of abstraction level, which indicates how the concept is rendered.

\textbf{Semantic Category Alignment.} To ensure interpretability, we align the creative image with a set of coarse semantic categories that serve as conceptual anchors in the shared CLIP embedding space.
After $\ell_2$-normalizing the visual and textual embeddings, we compute cosine similarity to measure semantic alignment:
\begin{equation}
z_c = \mathrm{cos}(\widetilde{\mathcal{F}}_v, \mathcal{F}_t),
\end{equation}
where the logits $z_c$ form a soft semantic alignment distribution, characterizing the content consistency and semantic uncertainty in creative transformations.

\textbf{Stylistic Abstraction Alignment.}
Beyond semantic content, creativity is also affected by \emph{how} an idea is visually expressed. In particular, the abstraction level indicates whether creative deviation arises primarily from stylistic transformation rather than semantic change. To capture this factor, we introduce a stylistic abstraction alignment branch to estimate the abstraction level as a proxy for stylistic deviation. 
More specifically, we predict abstraction from the feature norm $\|\widetilde{\mathcal{F}}_v\|$, motivated by evidence that the magnitude of CLIP image embeddings encodes informative cues that are suppressed by cosine similarity due to $\ell_2$ normalization~\cite{liao2025magnitude}. Empirically, we observe that the $\|\widetilde{\mathcal{F}}_v\|$ increases with the abstraction level (see Sec.~\ref{sec:feature_norm}). This suggests that $\|\widetilde{\mathcal{F}}_v\|$ provides a simple yet informative cue for stylization.
Accordingly, we apply a lightweight classification head to the creative image embedding:
\begin{equation}
z_s = \mathrm{MLP}(\|\widetilde{\mathcal{F}}_v \|),
\label{eqn:h}
\end{equation}
which outputs three abstraction categories: \emph{abstract}, \emph{general}, and \emph{concrete}. This branch provides auxiliary supervision during training and yields a stylistic signal used to regulate the allowable extent of creative deviation in subsequent hyperbolic modeling.

\subsection{Constrained Hyperbolic Representation}
The semantic and stylistic alignment signals specify \emph{where} creative deviation occur and \emph{how} an idea is visually expressed. However, they do not determine \emph{how far} a creative representation is allowed to deviate. Without explicit regulation, large deviations in latent space may correspond to arbitrary distortions rather than meaningful creative transformation. We therefore embed representations in a hyperbolic space and impose a style-conditioned constraint to regulate the allowable deviation range.

\textbf{Euclidean-to-Hyperbolic Projection.}
Given the creative visual embedding $\widetilde{\mathcal{F}}_v$, we map it into a $d$-dimensional Poincar\'e ball with curvature $\kappa>0$.. We first apply a learnable linear transformation $\mathbf{W}$, followed by the exponential map at the origin point $\mathbf{o}$: 
\begin{equation} 
\mathbf{h} = \exp^{\kappa}_{\mathbf{o}}(\mathbf{W}\widetilde{\mathcal{F}}_v).
\label{eqn: fvh}
\end{equation} 
In the Poincar\'e ball model, the norm $\|\mathbf{h}\|$ corresponds to the radial distance from the semantic center, providing a natural geometric measure of conceptual deviation: larger radial indicate greater departure from prototypical representations.

\paragraph{Style-conditioned Adaptive Constraint.}
Conceptual spaces are inherently non-uniform: concrete creations concentrate near dense cores, whereas abstract ones occupy sparser peripheral regions. Consequently, applying a uniform deviation constraint would either over-penalize abstract creations which legitimately require larger stylistic shifts, or under-constrain concrete depictions which should remain close to the semantic core for interpretability. To address this, we introduce an adaptive geometric constraint conditioned on the stylistic alignment signal, which modulates the allowable hyperbolic radius according to abstraction.  Specifically, the style logits $z_s$ are mapped to a continuous coefficient $\alpha \in (0,1)$ via a lightweight gating function:
\begin{equation}
\alpha = \sigma\big(g(z_s)\big),
\label{eqn:alpha}
\end{equation}
where $g(\cdot)$ denotes a learnable multilayer perceptron and $\sigma(\cdot)$ is the sigmoid function. The hyperbolic representation is then constrained as:
\begin{equation}
\|\mathbf{h}\| \le (1-\alpha)\frac{1}{\sqrt{\kappa}},
\end{equation}
which adaptively regulates the allowable deviation radius according to stylistic abstraction. The resulting constrained hyperbolic representations define a structured space in which novelty and appropriateness are subsequently evaluated.

\subsection{Creativity Scoring via Novelty-Appropriateness}
Inspired by the creativity theories~\cite{sternberg1999concept}, we model creativity through two coupled factors: \textit{novelty} measures how far the creative result deviates from its reference, whereas \textit{appropriateness} determines whether this deviation remains semantically coherent and thus constitutes meaningful creativity.

\textbf{Novelty Assessment.}
Given the hyperbolic embeddings $\mathbf{h}$ and $\widetilde{\mathbf{h}}$ for the original and creative images, we quantify novelty by the hyperbolic geodesic distance~\cite{poincare,ganea2018hyperbolic,hyper_CLIP, Pencils_to_Pixels}:
\begin{equation} D_{\mathrm{nov}} = d_{\mathbb{H}}(\mathbf{h}, \widetilde{\mathbf{h}}). 
\label{eqn:nov}
\end{equation} 
In the exponential metric expansion of hyperbolic space, this distance naturally captures large yet structured deviations.

\textbf{Appropriateness Assessment.}
Appropriateness is not an absolute property but depends on the degree of novelty: as deviation increases, the criterion for being ``reasonable'' shifts from literal similarity to conceptual interpretability. Therefore, we assess appropriateness by combining novelty with a measure of semantic coherence derived from semantic alignment. Specifically, we measure the semantic uncertainty from the content logits $z_c$:
\begin{equation}
H_{\mathrm{sem}} = - \sum_{c \in \mathcal{C}} p_c \log p_c,
\end{equation}
where $p_c = \mathrm{Softmax}(z_c)$. The lower uncertainty indicates concentrated semantic alignment, and higher uncertainty reflects semantic ambiguity. We then form a joint descriptor $\mathbf{u} = [D_{\mathrm{nov}}, H_{\mathrm{sem}}]$ and map it into hyperbolic space:
\begin{equation}
\mathbf{h}_a = \exp^{c}_{\mathbf{0}}(\mathbf{W}_a \mathbf{u}),
\end{equation}
with a learnable projection $\mathbf{W}_a$. As such, appropriateness is defined  as the signed distance to a hyperbolic hyperplane $\mathcal{H}$ parameterized by $(\mathbf{p}_0, \mathbf{w})$:
\begin{equation}
D_{\mathrm{app}} =
\mathrm{sign}\!\left(
\langle \mathbf{w}, \log_{\mathbf{p}_0}(\mathbf{h}_a) \rangle
\right)
\cdot
d_{\mathbb{H}}(\mathbf{h}_a, \mathcal{H}),
\label{eqn:dapp}
\end{equation}
which indicates whether the creative deviation falls within the region of meaningful conceptual expansion. For the Poincar\'e ball model~\cite{ganea2018hyperbolic}, this signed distance admits a closed-form expression.

\textbf{Creativity Assessment.} Finally, we map the appropriateness signal to a bounded creativity score:
\begin{equation}
S = \sigma(D_{\mathrm{app}} + b),
\end{equation}
where $b$ is a learnable bias term. This formulation ensures that creativity is rewarded only when novelty is justified by semantic coherence, penalizing both trivial reproductions and semantically incoherent deviations.

\section{Experiments}
\subsection{Experimental Settings}
To comprehensively assess both in-domain performance and cross-condition robustness, we conduct experiments under a unified evaluation protocol on the AuDrA benchmark~\cite{patterson2024audra}. This benchmark enables systematic analysis of creativity assessment models under controlled variations of rater composition and task formulation, allowing us to disentangle standard fitting ability from true generalization capability. The evaluations along three axes: (i) performance on held-out samples with consistent rater distributions, (ii) robustness to unseen annotators, and (iii) joint generalization to novel raters and task domains.

Model training and primary evaluation are performed on the \textbf{Primary Dataset (PD)}, which contains 11,075 abstract drawings annotated on a five-point creativity scale. The data are randomly partitioned into training, validation, and test splits with ratios of $70$\%, $10$\%, and $20$\%, respectively.  All reported in-domain results are measured on the held-out test split. Each sample is scored by $50$ trained annotators, resulting in highly consistent ratings with an intraclass correlation coefficient (ICC) exceeding 0.89, which provides a stable supervision signal for learning.

Beyond standard evaluation, we further assess generalization using three auxiliary subsets that are excluded from training. Two subsets, namely \textbf{Rater Generalization 1 (RG1)} and \textbf{Rater Generalization 2 (RG2)}, share the same abstract drawing domain as PD but are annotated by independent rater groups, enabling evaluation of robustness to rater shift. RG1 consists of $670$ samples rated by three new annotators (ICC = $0.73$), while RG2 contains $722$ samples rated by six new annotators (ICC = $0.90$). In addition, a more challenging subset, referred to as \textbf{Rater \& Task Generalization (FG)}, includes $679$ drawings depicting concrete object categories rather than abstract patterns, and is rated by three unseen annotators (ICC = $0.63$). This subset jointly probes both rater and task transferability. Statistical summaries of all subsets are provided in Table~\ref{tab:1}.

\begin{table}
\small
\centering
\caption{Summary of the AuDrA dataset~\cite{patterson2024audra}.}
\begin{tabular}{c c c >{\centering\arraybackslash}p{1.8cm} c}
\toprule
\textbf{Subset} & \textbf{Type} & \textbf{\#Samples} & \textbf{Raters} & \textbf{ICC} \\
\midrule
PD & Abstract & $11,075$ & $50$  & $>0.89$ \\
RG1 & Abstract & $670$ & $3$ new raters & $0.73$ \\
RG2 & Abstract & $722$ & $6$ new raters & $0.90$ \\
FG  & Objects & $679$ & $3$ new raters & $0.63$ \\
\bottomrule
\end{tabular}
\label{tab:1}
\vspace{-1em}
\end{table}

We follow the official training protocol of the AuDrA benchmark, with a batch size of $16$, a learning rate of $1\times10^{-5}$, and a maximum of $136$ training epochs. 
We adopt ViT-L/14 as the visual encoder and keep it frozen throughout training. All images are resized to $224\times224$. Model performance is evaluated using Spearman Rank Correlation Coefficient (SRCC) and Pearson Linear Correlation Coefficient (PLCC) between predicted creativity scores and the normalized Joint Rater Truth (JRT) annotations.

\subsection{Experimental results}

\textbf{Performance  on the Primary Set.}
We evaluate our HSCA on the primary test split against a diverse set of baselines, including  CLIP-based scoring models built upon ViT-B/32, ViT-B/16, ViT-L/14, RN50, and RN101 under the CLIPIQA+ formulation~\cite{wang2023exploring}, which scores images via similarity to predefined text prompts (``a photo of good creativity'' and ``a photo of bad creativity''). We further compare with recent creativity and image quality assessment (IQA) approaches, including AuDrA~\cite{patterson2024audra},  Pencils-to-Pixels~\cite{Pencils_to_Pixels}, and AGIQA~\cite{tang2025clip}. 
As shown in Table~\ref{tab2}, HSCA achieves the highest SRCC/PLCC of $0.87$/$0.88$, surpassing classical CNNs (VGG16~\cite{simonyan2014very}: $0.52$/$0.04$; ResNet50~\cite{he2016deep}: $0.78$/$0.79$), all CLIP-based variants including CLIP-RN101 (0.81/0.81), and prior creativity assessment models such as AuDrA ($0.80$/$0.79$) and Pencils-to-Pixels ($0.79$/-). These results suggest that HSCA more effectively captures the interplay of novelty and appropriateness in drawings, yielding predictions that better align with human judgments.

\begin{table}[ht]
\centering
\caption{Performance comparison on the primary test set.}
\label{tab2}
\setlength{\tabcolsep}{4pt}
\begin{tabular}{l l c c c}
\toprule
\textbf{Group} & \textbf{Model} & \textbf{Year} & \textbf{SRCC} & \textbf{PLCC} \\
\midrule
VGG & VGG16 & $2014$ & $0.52$ & $0.04$ \\
\midrule
\multirow{2}{*}{ResNet} 
    & ResNet34 & $2016$ & $0.79$ & $0.81$ \\
    & ResNet50 & $2016$ & $0.78$ & $0.79$ \\
\midrule
\multirow{5}{*}{CLIPIQA+}
    & CLIP (ViT-B/32) & $2023$ & $0.63$ & $0.65$ \\
    & CLIP (ViT-B/16) & $2023$ & $0.67$ & $0.68$ \\
    & CLIP (ViT-L/14) & $2023$ & $0.67$ & $0.69$ \\
    & CLIP (RN50)     & $2023$ & $0.80$ & $0.81$ \\
    & CLIP (RN101)    & $2023$ & $0.81$ & $0.81$ \\
\midrule
\multirow{3}{*}{Others} 
    & Audra           & $2024$ & $0.80$ & $0.79$ \\
    & AGIQA           & $2025$ & $0.79$ & $0.80$ \\
    & Pencils-to-Pixels & $2025$ & $0.79$ & - \\
\midrule
Ours & HSCA & - & $\textbf{0.87}$ & $\textbf{0.88}$ \\
\bottomrule
\end{tabular}
\vspace{-1em}
\end{table}

\begin{table*}[t]
\centering
\caption{Generalization performance on the three unseen subsets in terms of SRCC and PLCC.}
\label{tab3}
\setlength{\tabcolsep}{5pt}
\begin{tabular}{l l c cc cc cc}
\toprule
\textbf{Group} & \textbf{Model} & \textbf{Year} 
& \multicolumn{2}{c}{\textbf{RG1}} 
& \multicolumn{2}{c}{\textbf{RG2}} 
& \multicolumn{2}{c}{\textbf{FG}} \\
\cmidrule(lr){4-5} \cmidrule(lr){6-7} \cmidrule(lr){8-9}
& & & SRCC & PLCC & SRCC & PLCC & SRCC & PLCC \\
\midrule
VGG & VGG16 & $2014$ & $0.58$ & $0.56$ & $0.39$ & $0.32$ & $0.21$ & $0.12$ \\
\midrule
\multirow{2}{*}{ResNet}
    & ResNet34 & $2016$ & $0.78$ & $0.75$ & $0.68$ & $0.66$ & $0.45$ & $0.47$ \\
    & ResNet50 & $2016$ & $\textbf{0.80}$ & $\textbf{0.77}$ & $0.67$ & $0.65$ & $0.46$ & $0.47$ \\
\midrule
\multirow{5}{*}{CLIPIQA+}
    & CLIP (ViT-B/32) & $2023$ & $0.69$ & $0.65$ & $0.41$ & $0.43$ & $0.43$ & $0.44$ \\
    & CLIP (ViT-B/16) & $2023$ & $0.73$ & $0.70$ & $0.46$ & $0.47$ & $0.37$ & $0.39$ \\
    & CLIP (ViT-L/14) & $2023$ & $0.72$ & $0.69$ & $0.47$ & $0.48$ & $0.35$ & $0.36$ \\
    & CLIP (RN50)     & $2023$ & $0.77$ & $0.75$ & $0.67$ & $0.65$ & $0.43$ & $0.45$ \\
    & CLIP (RN101)    & $2023$ & $0.79$ & $0.76$ & $0.68$ & $0.66$ & $0.37$ & $0.38$ \\
\midrule
\multirow{2}{*}{Others}
    & AuDrA & $2024$ & $0.78$ & $0.75$ & $0.68$ & $0.67$ & $0.44$ & $0.46$ \\
    & AGIQA & $2025$ & $0.78$ & $\textbf{0.77}$ & $0.68$ & $0.67$ & $0.44$ & $0.45$ \\
\midrule
Ours & HSCA & $2025$ & $0.79$ & $\textbf{0.77}$ & $\textbf{0.72}$ & $\textbf{0.71}$ & $\textbf{0.55}$ & $\textbf{0.56}$ \\
\bottomrule
\end{tabular}
\end{table*}

\begin{table*}[htbp]
\centering
\caption{Ablation study results on PD, RG1, and FG. ``Nov.'' and ``App.'' denote novelty and appropriateness, respectively.}
\label{tab:ablation}
\setlength{\tabcolsep}{5pt}
\renewcommand{\arraystretch}{1.05}
\begin{tabular}{cccc|ccc}
\toprule
\multirow{2}{*}{\textbf{Nov.}} 
& \multirow{2}{*}{\textbf{App.}} 
& \multirow{2}{*}{\textbf{Adaptive Radius}} 
& \multirow{2}{*}{\textbf{Hyperbolic}}
& \textbf{PD} & \textbf{RG1} & \textbf{FG} \\
& & & & \textbf{SRCC\,/\,PLCC} & \textbf{SRCC\,/\,PLCC} & \textbf{SRCC\,/\,PLCC} \\
\midrule
\rowcolor{gray!10}
\cmark & \cmark & \cmark & \cmark &
$\textbf{0.87}\,/\,\textbf{0.88}$ &
$\textbf{0.79}\,/\,\textbf{0.77}$ &
$\textbf{0.55}\,/\,\textbf{0.56}$ \\

\xmark & \cmark & \cmark & \cmark &
${-0.01}\,/\,-0.01$ &
${-0.06}\,/\,-0.06$ &
$0.05\,/\,0.05$ \\

\rowcolor{gray!10}
\cmark & \xmark & \cmark & \cmark &
$0.86\,/\,0.85$ &
$0.78\,/\,0.76$ &
$-0.25\,/\, -0.20$ \\

\cmark & \cmark & \xmark & \cmark &
$0.84\,/\,0.85$ &
$0.78\,/\,0.76$ &
$0.49\,/\,0.51$ \\

\rowcolor{gray!10}
\cmark & \cmark & \xmark & \xmark &
$0.86\,/\,0.86$ &
$0.78\,/\,0.76$ &
$0.50\,/\,0.50$ \\
\bottomrule
\end{tabular}
\vspace{-1em}
\end{table*}

\textbf{Generalization on the other Three Sets.}
We further evaluate the cross-set generalization of different methods on RG1, RG2, and FG, and report the results in Table~\ref{tab3}. It is worth noting that reference images are unavailable in RG2; therefore, following a unified input protocol, we use a blank white image as the reference input.
Overall, HSCA consistently achieves the best performance across all three subsets, demonstrating strong robustness under both rater shifts and task variations. Specifically, on RG1, HSCA attains competitive SRCC/PLCC scores of $0.79$/$0.77$, comparable to the strongest CLIP-based and learning-based baselines.  More importantly, the performance gap becomes more evident on the more challenging RG2 and FG subsets, where HSCA surpasses all competing methods by clear margins. On RG2, HSCA improves PLCC from $0.67$ (Audra and AGIQA) to $0.71$, indicating better stability under increased annotation variance.
On FG, which involves cross-task generalization, HSCA achieves $0.55$/$0.56$ in SRCC/PLCC, outperforming the best baseline by over $10$\% relative improvement in correlation.
These results suggest that HSCA generalizes more reliably to unseen rating distributions and task settings, validating the effectiveness of our hierarchical creativity inference design for cross-domain deployment.

\begin{figure*}[t]
\centering
\includegraphics[width=0.96\linewidth]{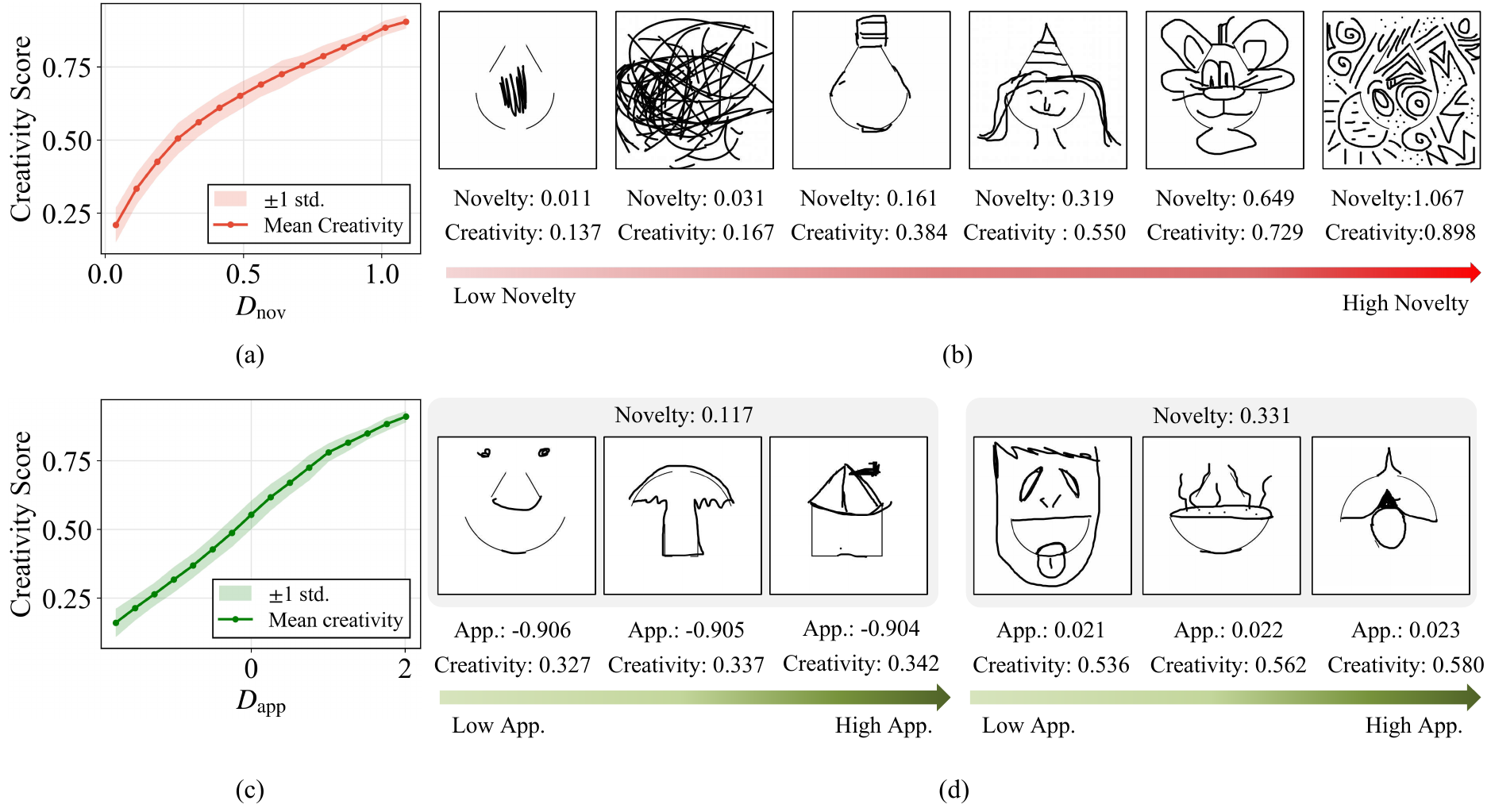}
\caption{\textbf{Correlation between creativity, novelty, and conditional appropriateness.}
(a) and (b): Correlations of novelty and appropriateness.  
(b) and (d): Example drawings illustrating how creativity and appropriateness tend to increase with novelty. Under similar novelty levels, higher conditional appropriateness leads to substantially higher creativity scores. ``App." denotes the appropriateness.}
\label{fig:hyperbolic_relation}
\end{figure*}

\begin{figure*}[htbp]
\centering
\includegraphics[width=0.9\linewidth]{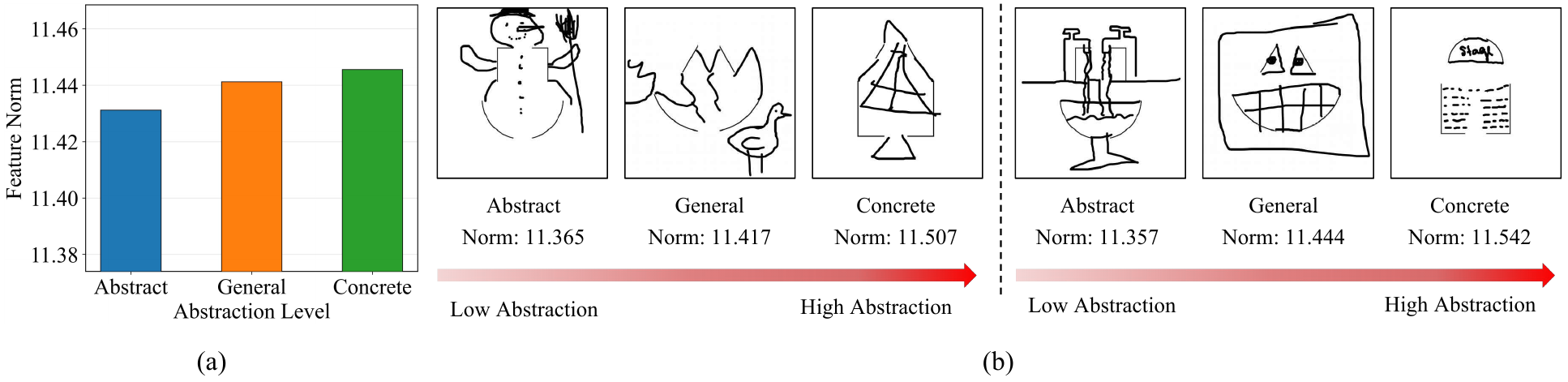}
\caption{\textbf{Representation magnitude versus abstraction level.}
(a) Mean norm of CLIP visual embeddings for drawings grouped by abstraction level. 
(b) Example drawings corresponding to each abstraction category. 
}
\label{fig:style_norm}
\vspace{-1em}
\end{figure*}
\subsection{Ablation Study}
We conduct ablation experiments to verify the necessity of each component in our framework. Specifically, the ablations aim to disentangle three key aspects of our design: (i) the dual-factor novelty--appropriateness formulation, (ii) the abstraction-conditioned adaptive radius constraint, and (iii) hyperbolic geometric modeling. 

\textbf{Novelty and Appropriateness Decomposition.}
To analyze the complementary roles of novelty and appropriateness, we independently ablate each component. 
Disabling the \emph{Novelty} branch removes explicit measurement of feature deviation between reference and drawing images, leaving only entropy-guided appropriateness estimation. This variant fails across all datasets, yielding near-zero or negative correlations (e.g., PD: $-0.01/-0.01$, RG1: $-0.06/-0.06$, FG: $0.05/0.05$), indicating that semantic uncertainty alone cannot capture creative variation.
In contrast, removing the \emph{Appropriateness} branch preserves novelty but discards conditional semantic entropy, replacing signed hyperbolic distances with direct regression of $D_{nov}$. While performance remains relatively stable on PD and RG1, it collapses on FG (from $0.55/0.56$ to $-0.25/-0.20$), suggesting that novelty without semantic conditioning easily degenerates into unstructured deviation that is poorly aligned with human creativity judgments. These results highlight that novelty provides the necessary stimulus for creativity, whereas appropriateness is essential for constraining novel deviations into meaningful creative outcomes.

\textbf{Effect of Adaptive Radius.}
Removing the abstraction-conditioned adaptive radius eliminates style-aware regulation on the allowable deviation range and enforces a fixed geometric boundary for all samples. As shown in Table~\ref{tab:ablation}, this leads to consistent and notable performance degradation across all three evaluation sets. The performance loss is particularly evident under generalization settings, indicating that a uniform constraint fails to accommodate varying abstraction levels in creative expressions. These results verify that dynamically adjusting the deviation radius according to abstraction cues is essential for maintaining stable creativity assessment across diverse styles and tasks.

\textbf{Role of Hyperbolic Geometry.}
Replacing hyperbolic modeling with standard Euclidean regression removes both geodesic novelty measurement and signed distance evaluation in hyperbolic space. While PD and RG1 remain relatively stable, FG performance degrades notably (from $0.55\,/\,0.56$ to $0.50\,/\,0.50$), indicating that hyperbolic geometry provides additional representational capacity for organizing large and hierarchical semantic deviations, which becomes particularly important under distribution shifts.

Overall, the full model consistently achieves the best performance across all subsets, with particularly strong gains on FG where semantic shifts and task variations are most pronounced. This demonstrates that integrating novelty and appropriateness through adaptive constraints in hyperbolic space is critical for robust creativity assessment under distribution changes.

\section{Visualization of Creativity Assessment}
\label{sec:feature_norm}
\textbf{Novelty and Conditional Appropriateness.}
We analyze how novelty (the hyperbolic geodesic distance $D_{\mathrm{nov}}$) and appropriateness (the signed distance $D_{\mathrm{app}}$ conditioned on the novelty) jointly relate to human creativity judgments in Fig.~\ref{fig:hyperbolic_relation}. 
Specifically, Fig.~\ref{fig:hyperbolic_relation}~(a) shows the relationship between creativity and novelty. Mean human creativity ratings with $\pm$1 standard deviation shading are plotted over $20$ equal-width bins, excluding bins with fewer than $10$ samples. Although creativity generally increases with novelty, the trend is clearly non-linear and less discriminative, indicating that deviation magnitude alone cannot distinguish meaningful creative transformations from arbitrary ones. To account for this, we further evaluate the conditional appropriateness in Fig.~\ref{fig:hyperbolic_relation}(c), which models whether a novel deviation remains structurally valid. Samples are binned using the same strategy as in (a). In contrast to novelty alone, creativity exhibits a near-monotonic and more linear increase with $D_{\mathrm{app}}$, indicating that once novelty is present, appropriateness becomes the dominant factor in determining perceived creativity.

Qualitative examples are presented in Fig.~\ref{fig:hyperbolic_relation}~(b) and (d). 
Fig.~\ref{fig:hyperbolic_relation}~(b) illustrates that human creativity ratings generally increase with novelty. 
In contrast, Fig.~\ref{fig:hyperbolic_relation}~(d) shows that {for drawings with comparable novelty, higher conditional appropriateness leads to substantially higher creativity scores}. These observations support our formulation that creativity is not determined by novelty in isolation, but by whether novel deviations remain semantically and structurally coherent.

\begin{figure}[htbp]
\centering
\includegraphics[width=\linewidth]{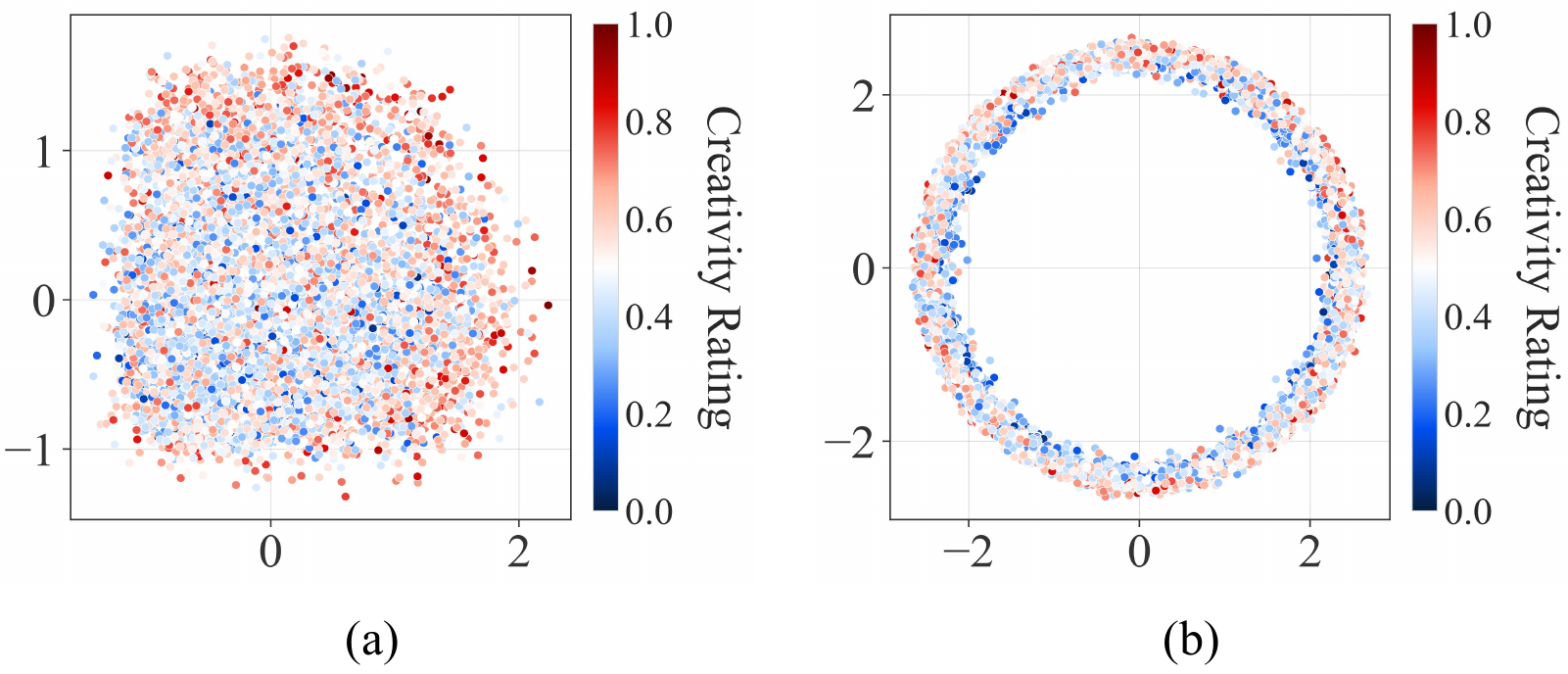}
\caption{\textbf{Feature geometry in Euclidean vs. hyperbolic space.}
(a) Euclidean features (b) corresponding hyperbolic embeddings.}
\label{fig:feature_geometry}
\end{figure}
\textbf{Abstraction Level via Feature Norm.}
To analyze how stylistic abstraction is reflected in visual representations, we visualize the norm of CLIP image embeddings for drawings grouped by abstraction level.

Fig.~\ref{fig:style_norm}~(a) reports the mean embedding norm for each style group, revealing that more abstract drawings tend to exhibit slightly larger representation norms. This suggests that abstraction influences how far samples deviate from the semantic center even before hyperbolic projection.

Fig.~\ref{fig:style_norm}~(b) shows representative examples from each style category. These results provide diagnostic evidence that abstraction information is implicitly encoded in the visual feature space, motivating our design of abstraction-conditioned adaptive constraints in the subsequent hyperbolic modeling stage. We emphasize that this analysis is intended to reveal representation tendencies rather than to claim large effect sizes.

\textbf{Euclidean \vs  Hyperbolic Feature Organization.}
To examine how hyperbolic projection reshapes the representation geometry, Fig.~\ref{fig:feature_geometry} visualizes both Euclidean and hyperbolic features of drawings from the primary test set using two-dimensional dimensionality reduction. For fair comparison, we visualize the penultimate-layer features used for creativity regression in both cases: (i) Euclidean features obtained after the MLP mapping from CLIP embeddings, and (ii) their corresponding hyperbolic projections in the Poincar\'e ball (\textit{i.e.},  $\mathbf{h}$ in Eqn.~(\ref{eqn:h})). Each point represents a drawing, colored by its ground-truth creativity rating.

As shown in Fig.~\ref{fig:feature_geometry}~(a), Euclidean features exhibit no clear global organization with respect to creativity scores, and samples with different ratings are largely intermingled. In contrast, Fig.~\ref{fig:feature_geometry}~(b) reveals a clear radial structure in hyperbolic space: higher creativity scores tend to concentrate farther from the center, while lower scores cluster closer to the origin. This hierarchical arrangement is consistent with the exponential volume growth property of hyperbolic geometry and supports our design that models creative deviation as increasing radial distance from semantic prototypes.

\vspace{1em}
\section{Conclusion}
We presented HSCA, a hyperbolic-space framework that models creativity as structured semantic deviation by jointly quantifying novelty and appropriateness. An abstraction-conditioned adaptive radius is further introduced to regulate semantic drift while preserving stylistic flexibility. Extensive experiments on AuDrA-Drawings demonstrate state-of-the-art alignment with human judgments and improved robustness under distribution shifts, with ablations validating the role of each component and the advantage of hyperbolic geometry. These results suggest hyperbolic representations offer a principled and interpretable basis for creativity assessment. Future work will extend this formulation to richer semantic conditioning and broader creative domains.

{
    \small
    \bibliographystyle{ieeenat_fullname}
    \bibliography{main}
}


\end{document}